\documentclass{article}

\usepackage[english]{babel}
\usepackage[margin=1in]{geometry}
\usepackage[utf8]{inputenc}
\usepackage[T1]{fontenc}
\usepackage{microtype}

\usepackage[utf8]{inputenc} 
\usepackage[T1]{fontenc}    
\usepackage{hyperref}       
\usepackage{booktabs}       
\usepackage{amsfonts}       
\usepackage{placeins}       
\usepackage{amsmath}        
\usepackage{nicefrac}       
\usepackage{microtype}      
\usepackage{lipsum}
\usepackage{graphicx}
\graphicspath{ {./images/} }

\usepackage{natbib}  
\usepackage{subcaption}
\usepackage{makecell}
\usepackage[table]{xcolor}
\usepackage[dvipsnames]{xcolor}  
\usepackage{xspace}
\newcommand{\fplat}{f_{\mathrm{plat}}}
\newcommand{\fproxy}{f_{\mathrm{proxy}}}
\newcommand{\fens}{f_{\mathrm{ens}}}
\newcommand{\Dpub}{\mathcal{D}_{\mathrm{pub}}}
\newcommand{\Dprox}{\mathcal{D}_{\mathrm{proxy}}}
\definecolor{todocolor}{RGB}{180,0,0}

\definecolor{darkgreen}{HTML}{006400}
\definecolor{darkblue}{rgb}{0, 0, 0.5}
\definecolor{LightGray}{gray}{0.9}
\definecolor{darkblue}{rgb}{0, 0, 0.5}

\newcommand{\cifarten}{CIFAR-10\xspace}
\newcommand{\cifarh}{CIFAR-100\xspace}
\newcommand{\fairface}{FairFace\xspace}
\newcommand{\resneteighteen}{ResNet-18\xspace}
\newcommand{\resnetfifty}{ResNet-50\xspace}
\newcommand{\regnet}{RegNet-Y-16GF\xspace}

\definecolor{figblue}{HTML}{00A6ED}
\definecolor{figred}{HTML}{F6511D}
\definecolor{figgreen}{HTML}{7FB800}

\title{Test-Time Collective Action:\\Proxy-Based Perturbations for Correcting Algorithmic Harms}

\usepackage{authblk}

\author[$\S$]{Meghana Bhange}
\author[$\S$]{Ulrich A\"ivodji}
\author[$\ddagger$]{Elliot Creager}
\affil[$\ddagger$]{University of Waterloo, Vector Institute}
\affil[$\S$]{\'ETS Montr\'eal, Mila - Quebec AI Institute}
\date{}

\begin{document}
\maketitle

\begin{abstract}
When machine learning systems under-perform for particular subgroups, affected users typically have no way to correct these disparities without relying on platform-level fixes. Existing approaches to algorithmic fairness rely on provider-centric approaches to correct these failures, leaving users with no external lever when faced with harm. Recent work in Algorithmic Collective Action shows that coordinated users can steer an algorithmic system toward a collective goal, but the existing mechanisms require the provider to retrain on the collective's modified data which users may not have control over. We propose \emph{Test-Time Collective Action (TTCA)}, a framework through which a group of users who share query access to the platform, can correct disparities affecting under-served subgroup without participating in the platform's training loop. We implement this through a proxy-based mechanism where the collective pools query access to a black-box API to extract a proxy of the platform, then optimizes a per-class universal perturbation against the proxy. Each member applies this perturbation to their own inputs at submission time, requiring no cooperation from the platform. We empirically evaluate the mechanism on \cifarten, \cifarh, and \fairface, showing that modestly-sized collectives close most of the subgroup accuracy gap, transfer across architectures (a small proxy can attack a larger platform), and improve worst-group accuracy, equal-opportunity gap, and disparate impact. A query-budget analysis comparing a per-user black-box attack baseline shows that pooling is cheaper than each subgroup member attacking alone. Test-time collective action thus offers corrective intervention to users when platform-side remediation is unavailable or delayed.

\begin{itemize}
    \raggedright
    \item \textbf{Code} - \url{https://github.com/tisl-lab/TestTimeCollectiveAction}
\end{itemize}

\end{abstract}

\section{Introduction}

\begin{figure*}[ht]
    \centering
    \includegraphics[width=0.90\linewidth]{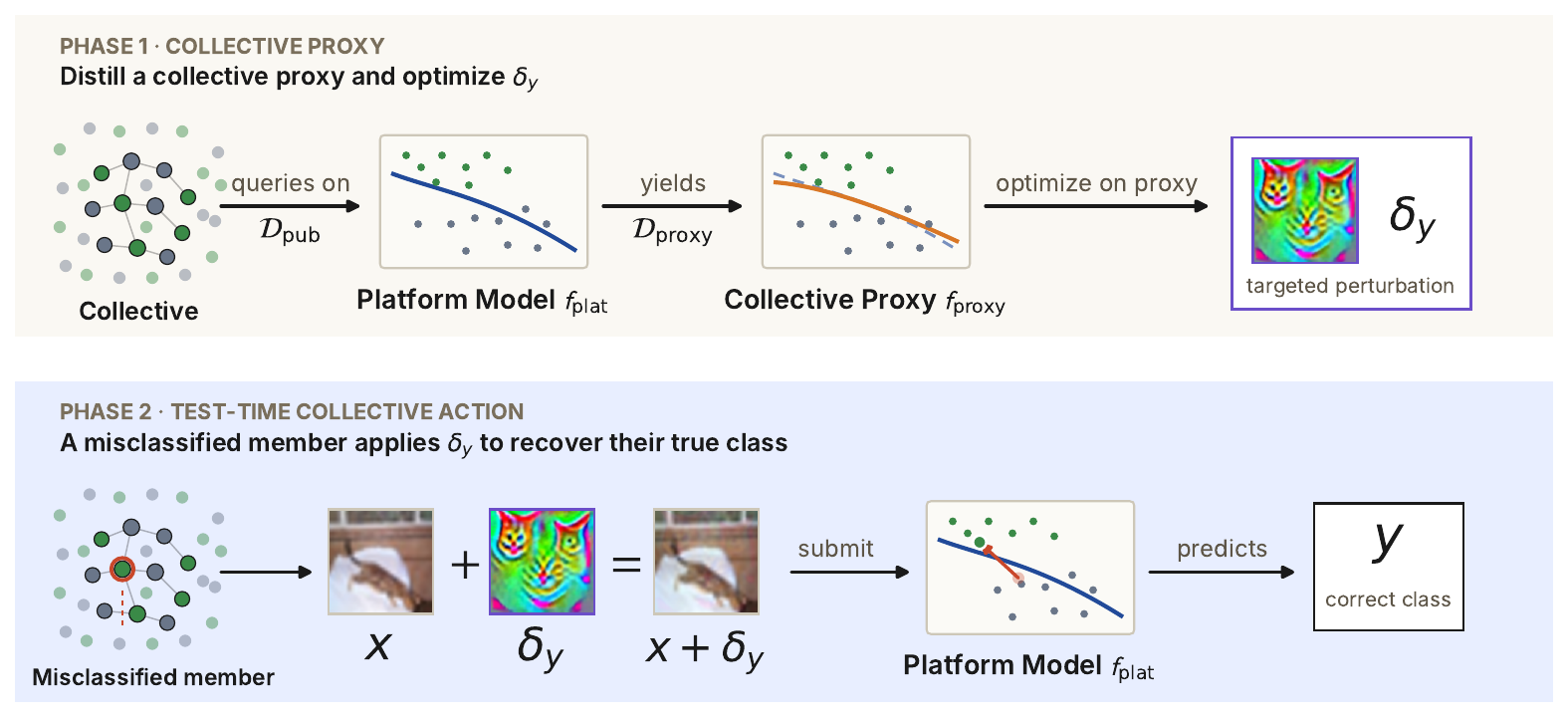}
    \caption{\emph{Test-Time Collective Action (TTCA)}. \emph{Top}: members pool their per-user query budgets to query the platform with a public image pool $\Dpub$, then train a collective proxy on the resulting query–response pool $\Dprox$, where the proxy boundary tries to approximate platform boundary. The collective optimizes a per-class perturbation $\delta_y$ against it. \emph{Bottom}: at test-time, each member of the under-served subgroup applies $\delta_y$ to their own input $x$, sending $x + \delta_y$ to the platform in place of $x$.}
    \label{fig:flow-diagram}
\end{figure*}

\begin{figure}[t]
  \centering
  \includegraphics[width=\columnwidth]{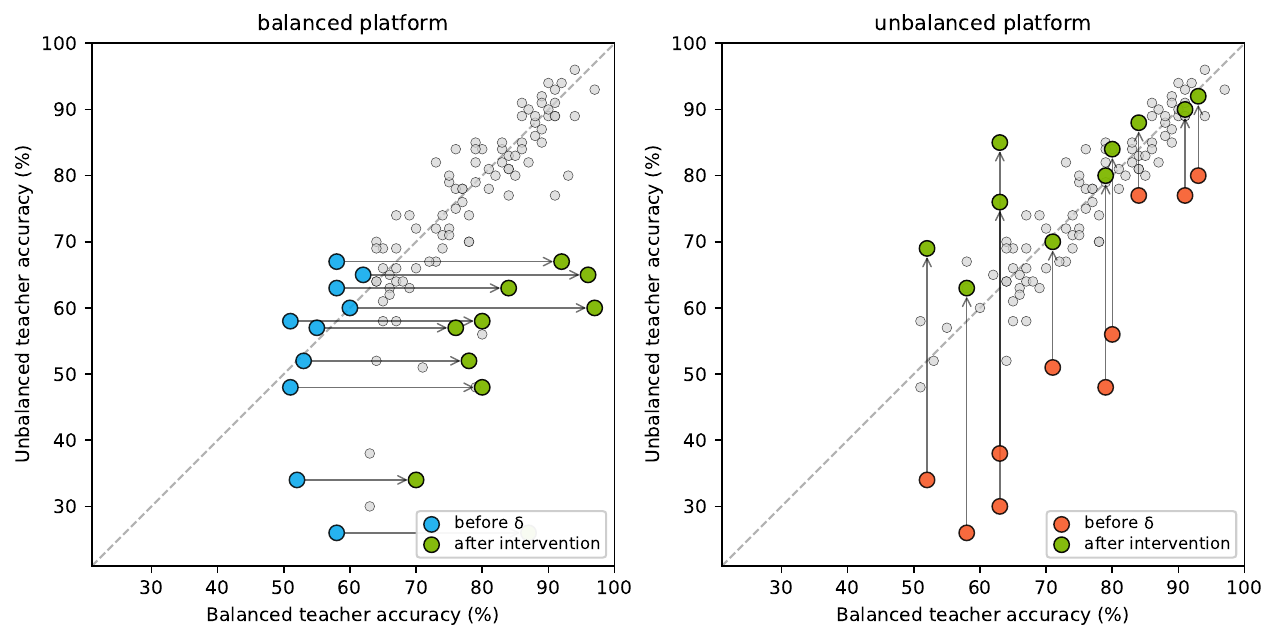}
   \caption{\emph{Per-class accuracy before and after the perturbation:} Test-time collective action helps the affected subgroup (defined here as bottom-10\% under-performing classes, but can alternatively include user demographics where available) for both balanced and unbalanced platform training regimes, without participating in the platform's training loop (subgroup selection and dataset regimes are defined in \S\ref{sec:data} and \S\ref{sec:harm-scenarios} respectively). The plot shows \cifarh classes plotted in balanced-platform subgroup accuracy, unbalanced-platform subgroup accuracy coordinates. \emph{Left}: bottom-10\% classes on the balanced platform (\textcolor{figblue}{$\bullet$}) move rightward after applying $\delta_y$, improving subgroup accuracy for balanced-platform regime. \emph{Right}: deliberately under-represented classes on the unbalanced platform regime (\textcolor{figred}{$\bullet$}) move upward after $\delta_y$. Other classes (gray) stay near the $y=x$ diagonal in both panels, so the perturbation acts as corrective recourse for the targeted subgroup rather than as a system-wide attack.}
  \label{fig:teacher-scatter}
\end{figure}

Machine learning systems can exhibit performance disparities across different user groups, with documented examples in computer vision~\citep{Buolamwini2018GenderShades}, safety features of language models~\citep{chehbouniRepresentationalHarmsQualityofService2024} and automated speech recognition~\citep{koeneckeRacialDisparitiesAutomated2020}. \citet{shelbySociotechnicalHarmsAlgorithmic2023} describe such failures as quality-of-service (QoS) harms. When users encounter these harms, provider-centric fairness interventions such as pre-processing, in-processing, and post-processing methods are commonly used to address them~\citep{catonFairnessMachineLearning2024}. This could be considered reasonable if platforms are willing to act towards reducing these harms; however, when remediation is unavailable, it leaves affected users with no lever of their own to address the situation. 

Recent work in Protective Optimization Technologies~\citep{kulynychPOTsProtectiveOptimization2020} and Algorithmic Collective Action~\citep{Hardt2024ACA} identifies promising avenues for user-side interventions that allow affected users to regain agency in machine learning systems. Algorithmic Collective Action \citep{Hardt2024ACA} shows how even a small, organized collective can have an out-sized effect in steering a platform's outputs in the collective's favor. However, these mechanisms require the platform to re-train on user contributions, a condition that users rarely have control over in real-world settings. \citet{radiya-dixit_data_2022} also discusses this limitation by showing that data poisoning-style interventions, as user-side protections, do not always work because providers can train on data collected before an attack was discovered, or simply wait for a more robust model before processing poisoned data. Hence, it is important for users to also have methods for taking corrective action at test time. In this work, we explore a test-time method that allows a group of users to pool their data contributions to generate a perturbation that mitigates the algorithmic harms of a frozen black-box model using only API query access. The high-level pipeline is shown in Figure~\ref{fig:flow-diagram}. which involves extracting a collective proxy from pooled queries, then applying the resulting perturbation at test-time.

\citet{shelbySociotechnicalHarmsAlgorithmic2023} define algorithmic harms as adverse experiences arising from the interaction between technical system components and societal power dynamics (we refer the reader to their survey for a full taxonomy on algorithmic harm). This work focuses on \textbf{quality-of-service (QoS) harms}, in which a system disproportionately under-performs for particular groups while maintaining strong performance for others. We focus on two regimes which could be the source of disparity: (i) a \textbf{balanced (or entire) platform}, where the platform sees the full training distribution and the under-served subgroup is whichever groups the model happens to perform worst on, and (ii) an \textbf{unbalanced platform}, where a controlled fraction of groups is deliberately under-sampled before training so that the under-served subgroup is the one the platform was systematically under-trained on. In both cases, the collective's goal is to close the accuracy gap on the affected subgroup without degrading performance for other groups.

In both settings, we assume that affected users are non-adversarial with respect to the platform's task objective and their goal is to attain the same quality of service that the rest of the population already receives. We assume scenarios in which the model is broadly useful to users but fails to deliver comparable performance or utility to a particular subgroup due to data imbalance or even deployment choices. Our framing thus treats this collective intervention as a form of a \emph{corrective action} for unequal outcomes when platform-side remediation is unavailable or delayed.

As an example of this setting, consider a user-driven photo-sharing platform deploys an image classifier that auto-tags every uploaded image and feeds those tags into its search and recommendation systems. The tool is broadly useful for users: most users' photos surface for the searches they should, reach an audience whose interests match. But for some users, the classifier produces systematically more errors than it does for the rest of the user base. The affected users do not want to break the search for the rest of the platform; they want their content surfaced as fairly as everyone else's. 

We study this setting empirically on standard image-classification benchmarks throughout this paper and evaluate the feasibility of such an intervention. Figure~\ref{fig:teacher-scatter} shows the intervention in action on \cifarh across both balanced and unbalanced platform training where a single per-class perturbation considerably improves the accuracy of the worst-off subgroup without disturbing the rest of the label space. 


In this work, we present Test-Time Collective Action as a mechanism to correct quality-of-service harms. We evaluate its performance on \cifarten, \cifarh, and \fairface under both balanced and unbalanced platform training. We show the following: 
\begin{enumerate}
    \item Subgroup accuracy improves even at the smallest pool size and once a proxy is faithful enough to the predictions of the platform model, additional queries do not significantly increase subgroup accuracy.  Indicating that a collective can generate an effective perturbation without exhausting their query budget or needing to label the full public dataset available to them. 
    \item Affected users attacking alone with per-image perturbations require up to 308× more total platform queries to reach comparable subgroup accuracy than a collective that pools queries once to train a proxy and generate reusable perturbations at B=700, narrowing to about 105× at the collective's accuracy ceiling (B=2500 and above). 
    Pooling can become cheaper than individual intervention with as few as five to ten contributing members.
    \item Lastly, the gains transfer to standard group-fairness metrics such as worst-group accuracy and equal-opportunity gap, which positions test-time collective action as an effective user-side fairness intervention. 
\end{enumerate}

\section{Related Works}
When users face the unwanted consequences of algorithmic systems and anticipate that platform-level remedies will not be available, the remaining lever is what users can do outside the system. \citet{kulynychPOTsProtectiveOptimization2020} propose \emph{Protective Optimization Technologies} (POTs) as an alternative to internal fixes by platform. POTs enable individuals or collectives to act from outside optimization systems rather than relying on platform cooperation. \emph{Data leverage} \citep{Vincent2021DataLeverage} refers to the power that individuals or groups have operating in data-driven technologies, dependent on their contributions. By withholding, redirecting, or altering data, users can affect the model's performance and outcomes. \citet{DeVrio2024ResponsesFromBelow} extend this lens to \emph{responses from below}, examining how people collectively build and use tools to resist or redirect algorithmic influence.

Algorithmic Collective Action (ACA)~\citep{Hardt2024ACA} formalizes how even a small, coordinated group of users can steer the algorithmic system in their favor by planting signals or erasing them. \citet{BenDov2025FairnessCollectiveAction} extend this to group fairness by implementing the erasure strategy, in which a minority-only collective relabels its own data to approximate counterfactual fairness. Recent work considers multiple simultaneous collectives~\citep{Karan2025AlgorithmicTwoCollectives} and robustness under differential privacy~\citep{solankiCrowdingOutNoise2025}. 

The current work in ACA assumes that the platform trains on the modified data of the collective; however, in practice, users may not always have control over the training loops of the platform. \citet{radiya-dixit_data_2022} discuss that once a perturbation is scraped, it cannot be revised, so the model trainer can simply wait for newer architectures that incidentally resist the perturbation, or adaptively train against the public attack tool. Test-time collective action alongside train-time action may give users more axes of agency in the system. An example of a deployment-time strategy explored in ACA literature is \emph{Decline Now}. \citet{sigg2025decline} model the \texttt{\#DeclineNow} campaign, in which DoorDash drivers collectively reject low-paying orders to trigger algorithmic price increases on the deployed allocation system.

Outside the ACA literature, evasion attacks \citep{biggioEvasionAttacksMachine2013} form the test-time counterpart to data-poisoning interventions. Instead of altering the data the platform trains on, the user perturbs the inputs at test-time to control the model's behavior. We re-purpose these mechanisms for correction purposes in this work (see the Adversarial Perturbations in \S\ref{sec:background} for details), showing that a small group of coordinated users can \emph{improve} model performance and fairness without requiring model retraining, or even knowledge of its architecture. 

Most directly related to our work is fairness-aware adversarial perturbation (FAAP)~\citep{wangFairnessawareAdversarialPerturbation2022}, which also uses a universal adversarial perturbation at test time to mitigate fairness disparities of a deployed model. The works overlap on the broad strategy but differ in their mechanisms and framing. FAAP trains a generator-discriminator pair on a deployed model to hide protected attribute information at test time. The motivation is that if the model cannot detect a sample's protected attribute, its predictions cannot correlate with that attribute. Our method takes a different approach, aiming to correct misclassifications of under-served groups using a universal perturbation that pushes the sample toward its true class, rather than censoring protected attributes. Our setting differs in both agency and access assumptions. FAAP studies fairness-aware perturbations as a bias-mitigation mechanism for deployed models, whereas TTCA frames perturbation generation as a user-side collective action problem under black-box API access. Affected users do not assume platform cooperation or internal model access; instead, they pool limited queries, extract a proxy, and generate reusable class-specific perturbations that can be applied at test time. Thus, TTCA contributes a collective, black-box, query-budgeted formulation of corrective perturbations rather than a new platform-side fairness intervention.

\section{Background}
\label{sec:background}
\paragraph{Black-box Model Extraction.}
Model extraction attacks exploit query access provided by prediction APIs to approximate, or ``steal", a deployed model~\citep{Tramer2016ModelStealingPredictionAPIs}. In a typical extraction attack, the adversary, who has black-box access, queries the target model on auxiliary inputs, collects the resulting query-response pairs, and uses them to train a surrogate model. Following prior work~\citet{jagielskiHighAccuracyHigh2020a}, the quality of the extraction can be measured using various metrics, such as accuracy (performance on the underlying learning task) and fidelity (matching the target model's predictions on a reference set). 
In this work, we use \texttt{Knockoff Nets}~\citep{orekondyKnockoffNetsStealing2018} to construct the collective proxy, since it assumes no knowledge of the train/test data used by the target model. Knockoff Nets offers two query strategies: a random strategy that uniformly samples from a public data pool, and an adaptive strategy that learns a sampling policy from auxiliary labels in that pool. We use the former to construct the collective’s proxy model.

\paragraph{Adversarial Perturbations.}\label{par:adv-pert}
Adversarial examples add small perturbations to an input, which can cause a model to change its output \citep{brunaIntriguingPropertiesNeural2013, Goodfellow2015AdversarialExamples}. Such perturbations belong to the class of evasion attacks~\citep{biggioEvasionAttacksMachine2013}, in which the adversary manipulates inputs at test time to induce erroneous predictions, without interfering with the model's training process. \citet{moosavi-dezfooliUniversalAdversarialPerturbations2017} introduced \emph{Universal Adversarial Perturbation} (UAP), which aims to find a single, universal perturbation that can misclassify images with high probability. Targeted variants of UAP, such as \emph{localized patches} and \emph{iterative tFGSM aggregation}, were subsequently developed by \citet{brownAdversarialPatch2018} and \citet{hiranoSimpleIterativeMethod2019}, respectively. In the context of per-image perturbation, query-only black-box attacks such as \emph{SimBA}~\citep{guoSimpleBlackboxAdversarial2019} fit a fresh perturbation to each image using only the platform's confidence scores. We use SimBA later as the individual baseline against which we compare the collective intervention.

Targeted UAPs are especially interesting in the context of collective user-side interventions because they can enable users to steer an algorithmic system toward a specific decision without accessing its training pipeline. Targeted UAP methods optimize a perturbation that moves inputs from non-target classes across the model's decision boundary and into the target class. However, when the perturbation is optimized against a proxy of the deployed model rather than the model itself, the proxy's decision boundary is an approximation of the platform's, and a perturbation tuned to cross the proxy's boundary may fail to cross the platform's. \citet{zengNarcissusPracticalCleanLabel2023} identify a closely related failure mode in the context of backdoor attacks, where surrogate models trained on out-of-distribution data, and propose instead optimizing the perturbation toward the target class itself rather than across its boundary. Our mechanism adopts this target-directed perturbation, but inverts both the threat model and the objective. The perturbation is applied at test time rather than during training (as in clean-label poisoning), and the intended predicted class is the input's ground-truth class. Thus, the goal is not to cause misclassifications but to steer the deployed platform model toward the correct decision.

\paragraph{Group fairness} 
Group fairness has been formalized along several axes to assess whether a classifier's outcomes are distributed equitably across groups. \emph{Disparate impact}~\citep{feldmanCertifyingRemovingDisparate2015} measures the rate at which an unprivileged group receives the favorable outcome compared to the rate for the privileged group. \emph{Equal opportunity}~\citep{hardtEqualityOpportunitySupervised2016} shifts the focus from outcome rates to error rates, requiring that true-positive rates be equalized across sensitive groups conditional on the advantaged outcome, so groups that genuinely qualify for a favorable label are recognized at equal rates. \emph{Worst-group accuracy} (WGA) ~\citep{sagawa*DistributionallyRobustNeural2019} measures the minimum classification accuracy across all groups, regardless of which group is ``privileged." Each metric captures a distinct fairness concern (rate of favor, error parity, worst-case service quality). The setting we study uses these metrics to determine whether the intervention improves group fairness in a corrective direction. The metrics are used to check if a crafted perturbation can reduce the existing disparity at inference time.

\section{Problem and Experimental Setup}
\label{sec:method}
\subsection{Threat model}
\label{sec:threat}

A platform operates a classifier $\fplat : \mathcal{X} \to [0,1]^{|\mathcal{Y}|}$ as a black-box API: users submit images $x \in \mathcal{X}$ and receive the platform's probability vector $\fplat(x)$ over the label space, whose argmax $\hat{y} = \arg\max_y \fplat(x)_y$ is the predicted class. Weights, architecture, training data, and intermediate representations are not exposed. The collective therefore observes only input--output behavior through the API and has no direct access to the model’s internal parameters, structure, training data, or hidden representations.

A \emph{collective} $C$ of $n$ users coordinates to correct misclassifications affecting an under-served subgroup $\mathcal{G}$. A collective $C$ may have members from $\mathcal{G}$ as well as other members in the system although may not necessarily overlap. Each member is constrained by a per-user query limit of $B$ requests, so the collective can pool a total budget of $Q = nB$ queries. Members share three resources: query access to $\fplat$ within their individual budgets; a public, unlabeled images pool $\Dpub$ that is disjoint from the platform's private training data; and the infrastructure to aggregate query results and distribute the generated perturbations. In our experiments, we vary the pooled budget $Q$ directly and report subgroup accuracy and fairness as functions of this collective budget. The decomposition $Q = nB$ is used only when we compare the collective to an individual user acting alone, where we hold $B$ fixed at each point of the sweep and vary $n$, so an individual at budget $B$ compares directly to a collective of $n$ users at the same per-user cost (total budget $nB$). Fixing $B$ rather than $Q$ isolates the effect of collective coordination as each user faces the same individual cost, so any improvement from cooperation is attributable to pooling rather than to one user having more queries available.

The collective's goal is to improve the platform's accuracy on the affected subgroup $\mathcal{G}$ at test time, without requiring the platform to retrain on modified inputs. The collective's lever is to add a small, bounded perturbation to the inputs of subgroup members before they are submitted to the API for prediction. Thus, this intervention repurposes the logic of evasion attacks toward a corrective end. Concretely, for each class $y$ represented in $\mathcal{G}$, the collective constructs a single perturbation $\delta_y$ with $\|\delta_y\|_\infty \le \varepsilon$ so that all subgroup members of class $y$ apply to their own inputs at test time. The perturbed input is $x' = \mathrm{clip}(x + \delta_y, 0, 1)$, where $\mathrm{clip}(z, 0, 1) = \min(\max(z, 0), 1)$ is applied element-wise to keep pixels in the valid range. The collective seeks to maximize subgroup accuracy:
\begin{equation}
  \mathrm{SubgroupAcc} \;=\; \Pr\!\bigl[\,\fplat(x') = y \;\big|\; y \in \mathcal{G}\,\bigr],
  \label{eq:subgroup_acc}
\end{equation}
The same $\delta_y$ has to work across every subgroup member of class $y$, which is what makes the perturbation \emph{universal} (where a single per-class perturbation, distributed once, is applied independently by each affected user).

\subsection{Mechanism}
\label{sec:mechanism}

Crafting $\delta_y$ would normally require access to gradients of the platform classifier, which are unavailable through a black-box API. The collective's workaround is to first train a surrogate classifier from the platform's query responses, and then optimize the perturbation against this surrogate.

\paragraph{(1) Query pooling.} First, each member uses their query budget to label a subset of $\Dpub$ with predictions from $\fplat$. These are aggregated into a shared pool $\Dprox = \{(x, \fplat(x))\}$ of $Q$ pairs, each linking an input $x$ to $\fplat$'s output rather than to a ground-truth label or one supplied by the collective. So the proxy trained on $\Dprox$ imitates $\fplat$'s predictions and even inherent biases rather than producing an accurate classifier of its own. This is the only stage that consumes platform queries; all subsequent steps are performed offline using $\Dprox$.

\paragraph{(2) Proxy extraction.}\label{part:mec-proxy-extraction} The collective then trains a surrogate $\fproxy$ on $\Dprox$ via KnockoffNets random sampling~\citep{orekondyKnockoffNetsStealing2018}, fitting a network with cross-entropy against the platform's soft-probability outputs. Because gradients on a single proxy can be noisy, we additionally train $K = 5$ proxies with independent random seeds and ensemble their outputs as $\fens = \tfrac{1}{K}\sum_{k=1}^K \fproxy^{(k)}$. Previous work has shown that ensembling multiple surrogate models can improve the transferability of crafted perturbations to target models~\citep{yaoUnderstandingModelEnsemble2025}.

\paragraph{(3) Perturbation optimization.} For each class $y$ in $\mathcal{G}$, the collective optimizes a perturbation $\delta_y$ against the proxy ensemble under an $\ell_\infty$ budget:
\begin{equation}
  \delta_y \;=\; \arg\min_{\|\delta\|_\infty \le \varepsilon}\; \mathbb{E}_{x \sim \Dprox^{y}}\!\bigl[\,\mathrm{CE}(\fens(\tilde{x}),\, y)\bigr],
  \label{eq:perturbation}
\end{equation}
where $\Dprox^{y}$ is the subset of queries the proxy labelled as $y$ and $\tilde{x} = \mathrm{clip}(x + \delta_y, 0, 1)$ is the perturbed input. A small total-variation regularizer ($10^{-4}$) and a Gaussian blur on $\delta$ at each step bias the perturbation toward low-frequency components. This design choice is motivated by prior findings on black-box transferability \citep{guoLowFrequencyAdversarial2020}. This formalization can be seen as a user-side counterpart to
boosting-style fairness corrections such as multiaccuracy \citep{KimGZ19} where the corrective signal is optimized against a
proxy ensemble from the under-served group's side and applied to the input at inference, rather than added by the platform to its own classifier during post-processing.

\paragraph{(4) Test-time application.} The collective distributes the set $\{\delta_y : y \in \mathcal{G}\}$ to its members. A member with input $x$ of class $y$ checks whether $y \in \mathcal{G}$; if so, they submit $\mathrm{clip}(x + \delta_y, 0, 1)$ instead of $x$; otherwise the input is sent unchanged. The pool $\Dprox$ is used for dual purposes. It is used for generating a training signal for \emph{Proxy extraction} in stage (2) and optimization set for \emph{Perturbation optimization} in stage (3), so no platform queries are spent beyond the initial $Q$.

\subsection{Datasets, partitions, and subgroups}
\label{sec:data}

We evaluate our method on different image-classification datasets. \cifarten and \cifarh \citep{krizhevsky2009learning} which are standard image-classification benchmarks of $32 \times 32$ color images with 10 and 100 classes respectively. For \cifarh we use the fine-grained labels rather than the 20 coarse super-class labels. In addition to the CIFAR datasets we also use \fairface~\citep{karkkainenfairface} which is a face-attribute dataset of 108{,}501 images, balanced across 7 race groups (White, Black, Indian, East Asian, Southeast Asian, Middle Eastern, Latino) and labelled with binary gender, race, and 9 age buckets. 
FairFace is used here because it gives us a real-world benchmark where the target label (gender) interacts with sensitive attributes (race, age), allowing us to evaluate the mechanism on demographic fairness rather than just per-class accuracy disparities. As the dataset is balanced across race/gender groups and unbalanced across age groups it also allows us to model the balanced and unbalanced platform regimes. We follow the dataset's binary gender annotation. However, gender recognition from faces is itself a contested practice \citep{Keyes18, ScheuermanPB19}, and we return to this scope question in \S\ref{sec:discussion}.

\paragraph{Disjoint platform/collective partitions.}
For each dataset, the official train split is shuffled with a fixed seed and divided $50{:}50$ (stratified by label) into a \emph{platform} partition $\mathcal{D}_P$ and a \emph{collective} partition $\mathcal{D}_C$. This is motivated by the assumption that the collective has no access to the platform’s training set, but can rely on auxiliary samples from the same underlying data distribution. This assumption reflects the fact that collective members are not external attackers probing an unrelated system; they are platform users whose inputs belong to the same operational environment as the model’s test-time inputs. Thus, the collective’s advantage comes from pooling user-side data and query access, not from privileged access to the platform’s training set or internal model state.
Within each partition, a further $90{:}10$ split yields train/val pools. The official test split is left intact, and neither party trains on it; it is only used for evaluation. 

\paragraph{Groups and subgroup definition.}
A \emph{group} is the unit at which fairness metrics are computed. On CIFAR datasets, the group is defined by the class label. On the \fairface dataset, the group is a (gender, sensitive attribute) pair, where the sensitive attribute is one of the 7 race buckets or one of the 9 age buckets tagged as attributes in the dataset. The under-served subgroup $\mathcal{G}$ is the bottom $10\%$ of groups by per-group accuracy on the shared test set.

\subsection{Balanced (or entire) vs.\ unbalanced platform training}
\label{sec:harm-scenarios}

We study QoS harm under two platform-training cases that differ in \emph{why} the under-served subgroup $\mathcal{G}$ is under-served. This distinction is important because the collective sees only the platform's outputs and not the reason behind them. An effective intervention should thus work whether the reason for the disparity comes from an incidental model weakness or from platform under-training on a subgroup due to a lack of data. The corrective objective from Equation~\ref{eq:subgroup_acc} is identical across both and only the construction of $\mathcal{G}$ differs.

\paragraph{Balanced (or entire) platform.}
The platform is trained on the full partition $\mathcal{D}_P^{\text{train}}$ with no deliberate sub-sampling. $\mathcal{G}$ is defined \emph{post hoc} as the bottom-$10\%$ groups by per-group accuracy on the shared test set. The collective's pool and the platform's data come from the same distribution, so the under-service arises from whatever combination of model weakness and natural data imbalance the dataset itself exhibits, rather than from any sub-sampling we have introduced.

\paragraph{Unbalanced platform.}
In this case, the platform is instead trained on a deliberately skewed variant of $\mathcal{D}_P^{,\text{train}}$. 10\% of groups are selected at random as under-represented and, within each, only a small fraction of examples is retained while the remaining groups are kept whole. The retention rate is set per dataset based on how aggressively the per-group pool can be cut without pushing the platform into underfitting: 10\% for \cifarten and \fairface, and 30\% for \cifarh, where the per-class platform pool is already small. Under-represented groups are picked uniformly at random for \cifarten/100 and \fairface (sensitive attribute: race); for \fairface (sensitive attribute: age) as the data is only balanced across race groups and not age buckets there is already a natural imbalance in the dataset. Thus, we instead pick the smallest age buckets to mirror the dataset's natural imbalance.

\subsection{Models and training parameters}
\label{sec:models}

\paragraph{Platform.} The platform backbones are \resneteighteen(\texttt{IMAGENET1K\_V1}), \resnetfifty(\texttt{IMAGENET1K\_V2})~\citep{ResNet} and \regnet(\texttt{IMAGENET1K\_V2})~\citep{RegNet}, pretrained on ImageNet, with only the final \texttt{fc} head replaced by a linear layer of size $|\mathcal{Y}|$. Both backbones are fine-tuned end-to-end on $\mathcal{D}_P^{\,\text{train}}$ with AdamW ($\eta = 10^{-3}$, weight decay $10^{-4}$), a cosine schedule over 30 epochs, batch size 128, and cross-entropy loss. We select the checkpoint with the highest validation accuracy.

\paragraph{Proxy.} The KnockoffNets surrogate uses the same backbone family as the platform by default. We use the implementation from the Adversarial Robustness Toolbox~\citep{nicolaeAdversarialRobustnessToolbox2019} where the implementation passes \texttt{use\_probability=True} to KnockoffNets, so the proxy distills the full output distribution rather than only the argmax label). We also test a cross-architecture setting in which a smaller \resneteighteen proxy is trained against a \resnetfifty and \regnet platform, modeling the case where users do not know the platform's architecture. Proxy hyper-parameters (AdamW, label smoothing $0.1$, 30 epochs, batch 128) are held fixed across configurations. We sweep the total query budget $Q$ at six log-spaced points from $500$ to $0.6\,|\mathcal{D}_C^{\,\text{train}}|$, and run five seeds per configuration so that subgroup-accuracy curves report mean $\pm$ standard deviation.

\paragraph{Perturbation.} Perturbation is optimized according to Stage (3) of mechanism mentioned in \S\ref{sec:mechanism}. Hyperparameter used for optimization are Adam ($\eta = 10^{-2}$, mini-batch 64, 200 iterations) with $\ell_\infty$ projection after each step. We use $\varepsilon = 0.031$ for CIFAR and $\varepsilon = 0.12$ for \fairface. 

\section{Experiments}

\begin{figure*}[!ht]
  \centering
  \begin{subfigure}[t]{0.49\textwidth}
    \centering
    \includegraphics[width=\linewidth]{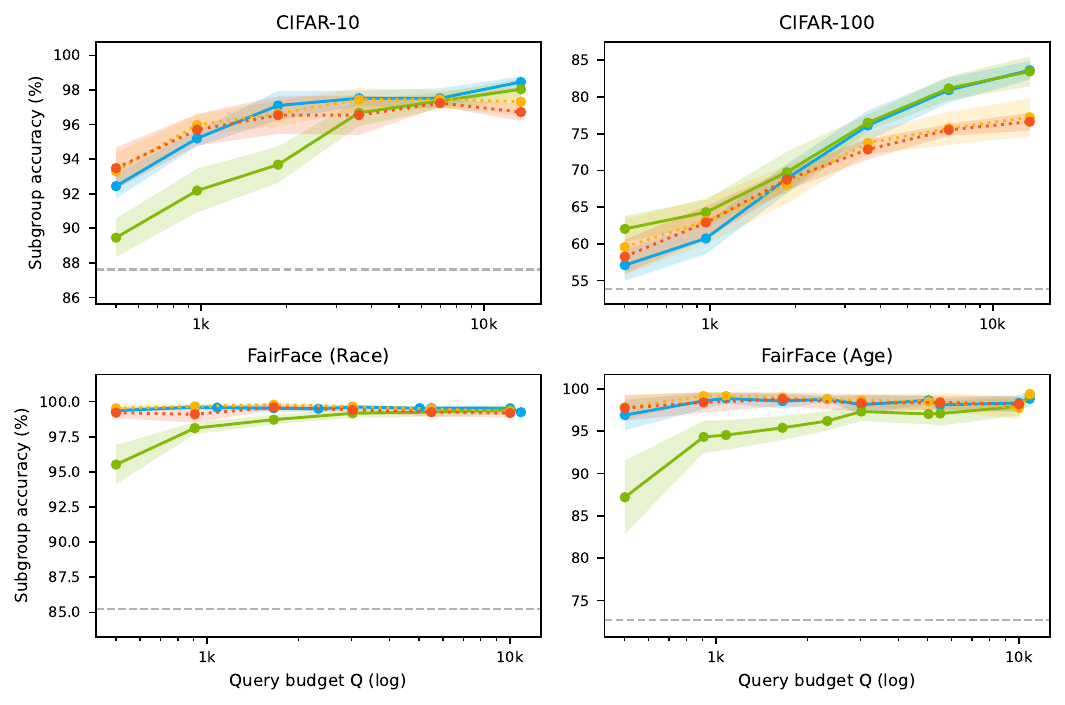}
    \caption{Balanced platform.}
    \label{fig:subgroup-balanced}
  \end{subfigure}
  \hfill
  \begin{subfigure}[t]{0.49\textwidth}
    \centering
    \includegraphics[width=\linewidth]{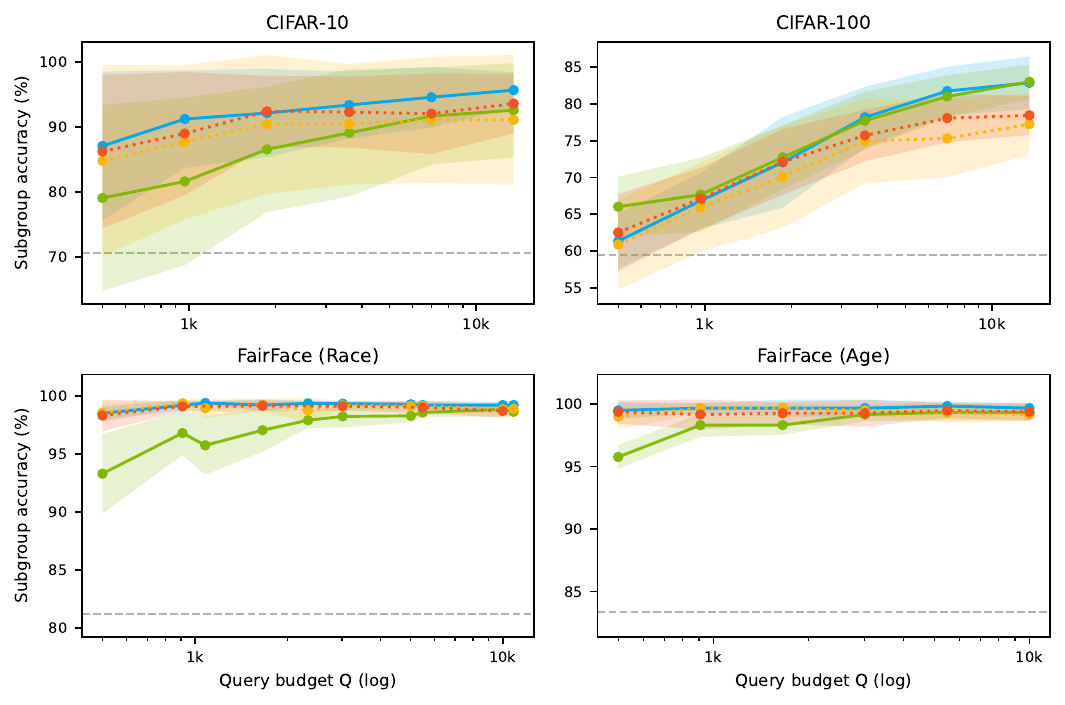}
    \caption{Unbalanced platform.}
    \label{fig:subgroup-unbalanced}
  \end{subfigure}
  \par\medskip
  \includegraphics[width=0.9\linewidth]{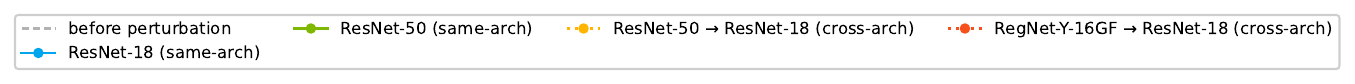}
  \caption{\emph{Subgroup accuracy after perturbation as a
function of pooled query budget $Q$ (log):} For \emph{balanced} and \emph{unbalanced} platform regimes even at the smallest query budget, the subgroup accuracy is above the pre-perturbation baseline for most of the configurations (dashed gray line) and the cross-architecture trajectories stay close to the same-architecture curves as $Q$ grows.}
  \label{fig:subgroup-combined}
  
\end{figure*}
\paragraph{Subgroup Accuracy and Fairness Metrics} The primary outcome we measure is the subgroup accuracy from Equation~\ref{eq:subgroup_acc} which is the platform's mean accuracy on the under-served subgroup $\mathcal{G}$ after members apply $\delta_y$ to their inputs at test time. We measure this as a function of the pooled query budget $Q$ across both platform-training regimes (\S\ref{sec:harm-scenarios}) and across platform backbones (\resneteighteen, \resnetfifty, \resnetfifty platform model $\rightarrow$ \resneteighteen proxy and \regnet platform model $\rightarrow$ \resneteighteen proxy), with five seeds per configuration. To measure if the intervention improves group fairness for $\mathcal{G}$ we also report the following metrics (\S\ref{sec:background}): worst-group accuracy, equal-opportunity gap, and disparate impact ratio. The collective's intervention is corrective if subgroup accuracy rises while these metrics improve or hold.
\paragraph{Individual baseline.} A natural alternative to collective pooling is for each affected user to attack the platform independently. In this baseline, each user runs a per-image targeted black-box attack on their own input, paying only for queries associated with that image and stopping once the platform predicts the correct class. This models an uncoordinated scenario in which users attempt to correct their own misclassifications without sharing data, queries, or perturbations with others.

We instantiate this baseline with targeted SimBA using the pixel variant~\citep{guoSimpleBlackboxAdversarial2019}. SimBA is appropriate for our setting because it requires only query access to the platform, the confidence score assigned to the target class, and the predicted label, which aligns with the type of API feedback assumed for the collective intervention. For each user, we run SimBA until the first successful correction or until the per-user query cap $B$ is reached. The realized query cost $q_i$ is therefore the actual number of queries spent by user $i$, capped at $B$, rather than simply the configured budget.

In the experiments, we sweep $B$, the per-user query budget that may be imposed by the platform through rate limiting. Both the individual and collective modes are evaluated on the same 1{,}000-user under-served subgroup selected in Section~\ref{sec:harm-scenarios}, using \cifarh across five seeds. We focus on \cifarh because it has the largest under-served subgroup among our benchmarks, making aggregate-cost scaling most visible.

For the individual baseline, each affected user attacks separately. Let $q_i=\min\{\tau_i,B\}$, where $\tau_i$ is the first query at which the individual attack succeeds for user $i$, and $B$ is the per-user query budget. If the attack does not succeed within budget, then $q_i=B$. The total query cost of the individual baseline is therefore $\sum_i q_i$.

For the collective intervention, users spend $N$ total queries to build the proxy model and generate the class-wise perturbations. Once constructed, the same perturbation $\delta_y$ can be applied by all 1{,}000 affected users in class $y$ at inference time without additional platform queries. To measure how many users are needed to build such a pool, we report the \emph{break-even collective size} $K^\star=\lceil N/B\rceil$, alongside the realized solo cost $\sum_i q_i$. Here, $K^\star$ is conservative: it is the minimum number of contributors required if each contributes their full budget $B$ to obtain a pool of size $N$ that matches the individual SimBA baseline at the same budget.

If more than $K^\star$ users participate, each contributor's share falls below $B$ while the pool size remains $N$, reducing the per-user query cost. Alternatively, if additional members each contribute up to $B$, the collective can build a larger pool and potentially improve accuracy beyond the individual baseline. The remaining $1{,}000-K^\star$ subgroup users can still benefit from $\delta_y$ without contributing any queries. Because the perturbation is reusable for the current deployed model, future affected users in class $y$ who encounter the same failure mode can also apply it directly rather than generating a new fix from scratch.

\section{Results}
The experiments show that test-time collective action can be an effective corrective intervention. The result is consistent across \cifarten, \cifarh, and \fairface; under both balanced and deliberately skewed platform-training regimes; and even when the collective's proxy is smaller than the platform it queries.

\subsection{Subgroup accuracy improves at low query budgets}
\label{sec:results-subgroup}

\begin{figure}[t]
  \centering
  \begin{subfigure}[t]{0.40\columnwidth}
    \centering
    \includegraphics[width=\linewidth]{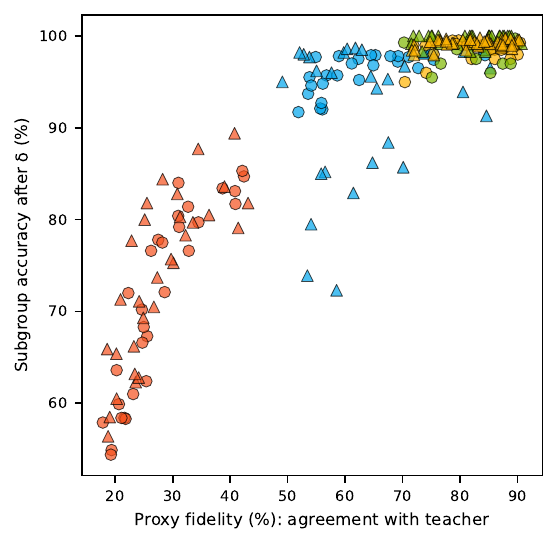}\\[2pt]
    \includegraphics[width=\linewidth]{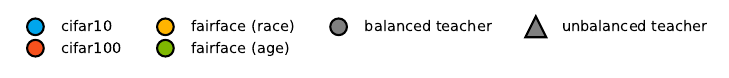}
    \caption{\emph{Proxy fidelity vs subgroup accuracy after $\delta_y$:} For CIFAR-10 and FairFace, subgroup accuracy saturates well below highest fidelity; however, CIFAR-100 remains in the climbing region and continues to benefit from higher $Q$. This indicates a faithful enough proxy suffices for most datasets, so users do not always need to exhaust their entire query budgets.}
    \label{fig:proxy-fidelity-vs-acc}
  \end{subfigure}
  \hfill
  \begin{subfigure}[t]{0.56\columnwidth}
    \centering
    \includegraphics[width=\linewidth]{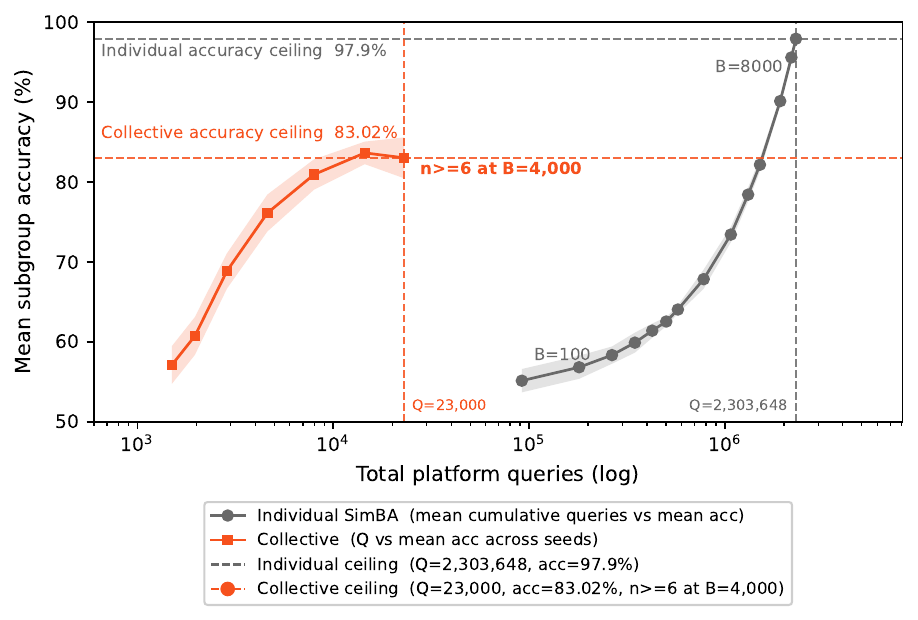}
    \caption{\emph{Subgroup Accuracy vs Total Platform queries (log):} Test-time collective action requires far fewer platform queries than individuals attacking alone however has a lower accuracy ceiling. Individual SimBA (gray) attains 97.9\% only after $\approx$2.3M cumulative queries from the 1,000-user subgroup. In contrast collective spends $N=22{,}000$ queries once to train a shared proxy, then derives a single universal perturbation $\delta$ that every member applies at inference, so the platform-query requests stays flat at $Q = 23{,}000$ regardless of how many of the 1,000 users participate; break-even at $n \ge 6$ attackers each willing to spend $B=4{,}000$ queries individually. However, even when the collective consumes the entire available public pool, accuracy ceilings at 83.02\% which is below the individual ceiling.}
    \label{fig:individual_vs_collective_cifar100_cost}
  \end{subfigure}
  \caption{Proxy fidelity and individual-vs-collective cost analyses.}
  \label{fig:proxy-and-cost}
\end{figure}

\paragraph{Subgroup Accuracy improvements} Figures~\ref{fig:subgroup-balanced} and~\ref{fig:subgroup-unbalanced} show that the mechanism works at even the smallest query budgets across all three benchmarks and both platform-training regimes. We report subgroup accuracy after applying the collective perturbation as a function of the pooled query budget $Q$, distinguishing between platforms trained on a balanced corpus (Figure~\ref{fig:subgroup-balanced}) and those trained on deliberately skewed data (Figure~\ref{fig:subgroup-unbalanced}). In most panels, the post-perturbation curves exceed the pre-perturbation baseline (dashed gray line) even at the smallest query budgets. This indicates that relatively small query pools are sufficient to improve subgroup accuracy. Appendix~\ref{app:white-image-test} further probes what $\delta_y$ encodes in isolation by applying the perturbation to a blank white image. The results reveal a class-specific signal already at $N=500$, suggesting that the collective can produce an effective perturbation even with a modest query budget. These findings show that the intervention benefits affected users without requiring the collective to exhaust its available queries. Higher pooled query budgets improve subgroup accuracy across all datasets until performance approaches its ceiling. Platforms trained on deliberately skewed data exhibit substantially higher seed-to-seed variance than those trained on balanced data. This variance is likely driven by the random selection of underrepresented groups across seeds. Consistent with this explanation, variance is much lower in the deterministic \fairface's by-age setting, where we select the smallest naturally occurring age buckets rather than sampling the affected groups at random.

\paragraph{Cross-architecture transfer} The collective may rely on a smaller proxy model when the platform architecture is unknown or when users face computational constraints. Figure~\ref{fig:subgroup-combined} also evaluates cross-architecture transfer by training the proxy as a \resneteighteen while the platform uses a \resnetfifty or \regnet. Across all datasets, the cross-architecture trajectory closely follows the same-architecture setting, indicating that perturbations crafted on a smaller \resneteighteen proxy transfer effectively to the larger \resnetfifty or \regnet platform. 

\paragraph{Proxy fidelity and Subgroup Accuracy} Effective perturbations do not always require a high-fidelity proxy. Figure~\ref{fig:proxy-fidelity-vs-acc} plots the proxy's fidelity against the subgroup accuracy the resulting perturbation achieves. Markers cluster tightly along an upward trend across every dataset. Hence, a larger $Q$ leads to a more faithful proxy, and a more faithful proxy results in a perturbation that transfers back to the platform more reliably. On \cifarten and \fairface, where the proxy reaches $60$-$90$\% fidelity within the budgets we test, the accuracy curve flattens near ceiling. \cifarh, however it never exceeds $\sim$42\% fidelity in our sweep and is still climbing, so additional queries continue to pay off. Here we can see once marginal fidelity stops translating into marginal accuracy, members can stop pooling rather than exhausting their per-user budgets or labeling the full public pool.

\subsection{Individual vs Collective Action}
\label{sec:ind-vs-coll}

\begin{table}[t]
  \centering
  \caption{\emph{Query-budget analysis on \cifarh:} At every per-user budget $B$ below the collective\textquotesingle s $\sim$83\% accuracy ceiling, pooling matches the individual SimBA baseline with far fewer platform queries, and the margin grows as $B$ is higher. The highlighted row is the largest $B$ at which a collective of $K^\star$ members still matches individual; the rows below it are budgets where individual surpasses the collective ceiling (by using two orders of magnitude more queries), and no pool size, including the full $\mathcal{D}_{\mathrm{pub}}$, closes the gap.}
  \label{tab:individual-vs-collective-cifar100}
  \vspace{2em}
  \scriptsize
  \setlength{\tabcolsep}{2.0pt}
  \renewcommand{\arraystretch}{1.15}
  \begin{tabular}{@{}ccccccc@{}}
    \toprule
      \makecell[c]{Query \\Budget\\($B$)} &
      \makecell[c]{Individual\\acc.\,(\%)} &
      \makecell[c]{Collective\\acc.\,(\%)} &
      \makecell[c]{Total\\ Collective\\Query Pool ($N$)} &
      \makecell[c]{Collective\\Members\\ ($K^\star$)} &
      \makecell[c]{Avg. \\Individual\\queries ($\overline q$)} &
      \makecell[c]{Total \\Individual\\queries ($\sum q_i$)} \\
    \midrule
    100 & 55.1{\tiny\,\textpm\,1.5} & 57.1{\tiny\,\textpm\,2.4} & 500{\tiny\,\textpm\,0} & 5 & 92{\tiny\,\textpm\,3} & 91{,}817{\tiny\,\textpm\,2{,}719} \\
    200 & 56.8{\tiny\,\textpm\,1.4} & 59.2{\tiny\,\textpm\,1.2} & 780{\tiny\,\textpm\,256} & 4 & 180{\tiny\,\textpm\,6} & 180{,}050{\tiny\,\textpm\,5{,}703} \\
    300 & 58.3{\tiny\,\textpm\,1.1} & 60.2{\tiny\,\textpm\,2.2} & 874{\tiny\,\textpm\,209} & 4 & 265{\tiny\,\textpm\,8} & 264{,}973{\tiny\,\textpm\,8{,}326} \\
    400 & 59.9{\tiny\,\textpm\,1.3} & 65.9{\tiny\,\textpm\,4.5} & 1{,}415{\tiny\,\textpm\,643} & 4 & 347{\tiny\,\textpm\,10} & 346{,}875{\tiny\,\textpm\,10{,}471} \\
    500 & 61.4{\tiny\,\textpm\,0.9} & 66.4{\tiny\,\textpm\,3.7} & 1{,}508{\tiny\,\textpm\,494} & 3 & 425{\tiny\,\textpm\,12} & 425{,}414{\tiny\,\textpm\,12{,}460} \\
    600 & 62.5{\tiny\,\textpm\,0.6} & 67.9{\tiny\,\textpm\,3.3} & 1{,}689{\tiny\,\textpm\,403} & 4 & 502{\tiny\,\textpm\,14} & 501{,}548{\tiny\,\textpm\,13{,}812} \\
    700 & 64.0{\tiny\,\textpm\,0.5} & 68.9{\tiny\,\textpm\,2.2} & 1{,}869{\tiny\,\textpm\,0} & 3 & 575{\tiny\,\textpm\,15} & 574{,}971{\tiny\,\textpm\,14{,}542} \\
    1{,}000 & 67.8{\tiny\,\textpm\,1.2} & 72.1{\tiny\,\textpm\,3.6} & 2{,}566{\tiny\,\textpm\,955} & 3 & 779{\tiny\,\textpm\,18} & 778{,}518{\tiny\,\textpm\,17{,}620} \\
    1{,}500 & 73.4{\tiny\,\textpm\,1.0} & 77.5{\tiny\,\textpm\,1.1} & 4{,}286{\tiny\,\textpm\,1{,}508} & 3 & 1{,}072{\tiny\,\textpm\,25} & 1{,}071{,}639{\tiny\,\textpm\,25{,}140} \\
    2{,}000 & 78.4{\tiny\,\textpm\,0.9} & 80.2{\tiny\,\textpm\,2.3} & 6{,}309{\tiny\,\textpm\,1{,}508} & 4 & 1{,}313{\tiny\,\textpm\,33} & 1{,}313{,}186{\tiny\,\textpm\,33{,}334} \\
    \rowcolor{green!15!gray!10} \textbf{2{,}500} & \textbf{82.2{\tiny\,\textpm\,0.8}} & \textbf{82.3{\tiny\,\textpm\,1.5}} & \textbf{14{,}293{\tiny\,\textpm\,7{,}522}} & \textbf{6} & \textbf{1{,}509{\tiny\,\textpm\,42}} & \textbf{1{,}508{,}935{\tiny\,\textpm\,42{,}121}} \\
    \midrule
     4{,}000 & 90.1{\tiny\,\textpm\,0.5} & 83.0{\tiny\,\textpm\,2.6} & 22{,}000{\tiny\,\textpm\,0} & 6 & 1{,}915{\tiny\,\textpm\,57} & 1{,}914{,}676{\tiny\,\textpm\,56{,}880} \\
    6{,}000 & 95.6{\tiny\,\textpm\,0.6} & 83.0{\tiny\,\textpm\,2.6} & 22{,}000{\tiny\,\textpm\,0} & 4 & 2{,}184{\tiny\,\textpm\,69} & 2{,}183{,}551{\tiny\,\textpm\,69{,}278} \\
     8{,}000 & 97.9{\tiny\,\textpm\,0.2} & 83.0{\tiny\,\textpm\,2.6} & 22{,}000{\tiny\,\textpm\,0} & 3 & 2{,}304{\tiny\,\textpm\,80} & 2{,}303{,}648{\tiny\,\textpm\,80{,}329} \\
    \bottomrule
  \end{tabular}
\end{table}

\paragraph{Subgroup-accuracy break-even occurs with only a few members.}
We sweep the per-user query budget $B$ and compare the individual SimBA baseline to the collective intervention on \cifarh across seeds. Table~\ref{tab:individual-vs-collective-cifar100} reports the individual baseline's mean subgroup accuracy, the collective accuracy achieved by the smallest matching query pool, the corresponding pool size $N$, the break-even collective size $K^\star=\lceil N/B\rceil$, and the realized cumulative solo cost $\sum_i q_i$.

At $B=100$, individual SimBA reaches only 55.1\% subgroup accuracy, with an average cost of approximately $92$ queries per user. This average reflects both users whose predictions are corrected quickly and users who exhaust the 100-query cap without success. In contrast, the collective reaches higher subgroup accuracy, 57.1\%, with only $N=500$ total query-response pairs. This corresponds to $K^\star=5$, meaning that five users each contributing their full 100-query budget are sufficient to build a pool that outperforms the individual baseline. Across all budgets up to the collective's accuracy ceiling, $K^\star$ remains in the single digits (ranging from three to six), indicating that only a small number of contributors is needed for pooling to become competitive in subgroup accuracy.

\paragraph{Pooling also reduces the average per-contributor cost.}
The cost-sharing calculation also favors pooling once the collective is slightly larger than $K^\star$. With $K$ contributors, the accounting cost per contributor is $N/K$. This falls below the average solo cost once
\[
K \ge \left\lceil \frac{N}{\overline q} \right\rceil,
\qquad
\text{where }
\overline q=\frac{1}{|\mathcal{G}|}\sum_{i\in\mathcal{G}} q_i .
\]
At $B=100$, this threshold is $\lceil 500/92\rceil=6$, only one more than the subgroup-accuracy break-even. This is a cost-sharing calculation rather than an additional experiment over all six-user subsets: it means that if six contributors contribute the same $N=500$ query pool, their average contribution is lower than the average number of queries spent by a user attacking alone. Near the collective's accuracy ceiling, the threshold remains modest. At $B=2{,}500$, the collective requires approximately $N=14{,}293$ queries, while the average solo cost is about $1{,}509$ queries, so roughly ten contributors are enough for the average per-contributor share to fall below the average solo cost.

This comparison highlights the main advantage of pooling. A targeted black-box attack can correct an individual user's prediction when that user has enough query budget, but the cost is not amortized: the next user facing the same failure must run a new per-image attack and pay a separate query cost. By contrast, the collective pays the proxy-extraction and perturbation-generation cost once, amortizes it across contributors, and produces a reusable perturbation $\delta_y$ that can be applied by non-contributing affected users at test time. This is especially valuable when many users face the same failure mode, or when individual users cannot sustain the solo attack loop because of account-level rate limits or other query constraints. The benefit persists for future affected users as long as the deployed model remains unchanged.

\paragraph{Well-resourced individuals can exceed the collective subgroup accuracy.}
The comparison also reveals a trade-off between individual and collective intervention, as shown in Figure~\ref{fig:individual_vs_collective_cifar100_cost}. The individual baseline reaches a higher subgroup accuracy on \cifarh, 97.9\%, than the collective, which saturates around 83.0\%. This is expected: SimBA optimizes a separate perturbation for each input, using the platform's own scores on that specific image, and can continue querying until the prediction crosses the decision boundary or the budget is exhausted. By contrast, the collective learns a universal perturbation $\delta_{y^\star}$ that must work for all affected subgroup members of class $y^\star$.

Above the collective's accuracy ceiling, no break-even pool size is defined because the present universal-perturbation mechanism cannot reach the individual baseline's accuracy, even when the available public pool is exhausted. However, this higher individual accuracy comes at a much larger aggregate query cost. The collective reaches its ceiling with roughly $23{,}000$ total platform queries paid once and reused across users, whereas bringing the 1{,}000 subgroup users to the individual ceiling requires approximately $2.3$ million cumulative queries. Thus, well-resourced individuals can obtain higher accuracy by attacking alone, but collective pooling remains far more query-efficient at the subgroup level.

\subsection{Fairness}

\begin{figure}[htbp]
  \centering
  \includegraphics[width=\columnwidth]{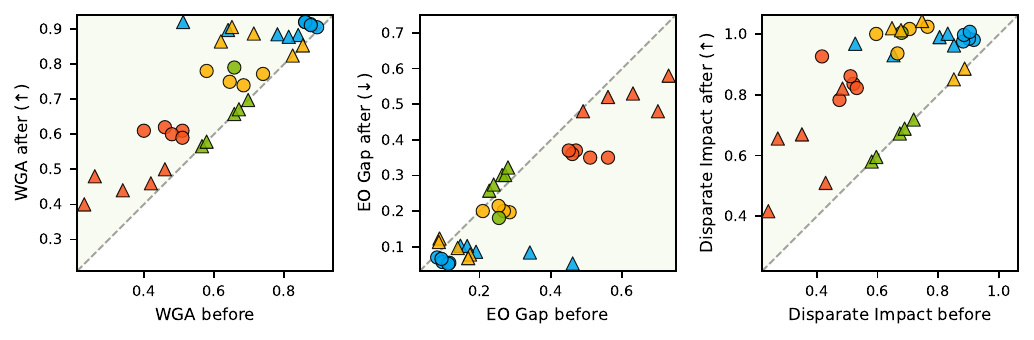}\\[2pt]
  \includegraphics[width=0.95\columnwidth]{figures/legend_lift.pdf}
  \caption{%
    \emph{Fairness metrics before vs.\ after the collective perturbation:} Each marker is one (dataset, sensitive attribute, platform-training regime) configuration at the smallest pool size that reaches our 75\% target accuracy; the shaded half-plane in each panel is the improving direction. Most markers fall on the improving side across all three metrics.}
  \label{fig:fairness-r18}
\end{figure}

\label{sec:results-fairness}

\paragraph{Subgroup-accuracy gains translate into improvements on group-fairness metrics.}
Figure~\ref{fig:fairness-r18} compares worst-group accuracy (WGA), equal-opportunity gap (EOD), and disparate impact ratio (DI) before and after applying the collective perturbation. Each marker corresponds to one experimental setting, defined by a dataset, sensitive attribute, and platform-training regime, evaluated at the smallest pool size that reaches the 75\% subgroup-accuracy target. We use this 75\% target as a common reference point: it is high enough to indicate that the perturbation is effective, but low enough to be reached across all settings. This avoids confounding the fairness comparison with differences in accuracy ceilings across datasets and regimes.

Most settings move in the improving direction for all three metrics: above the $y{=}x$ diagonal for WGA and DI, and below it for EOD. WGA shows the largest and most consistent improvement, which is expected because it is the metric most directly tied to the under-served subgroup targeted by the perturbation. In particular, settings that initially have WGA in the 0.3--0.4 range typically improve beyond 0.4, and several \cifarten settings reach 0.85 or higher after the intervention.

This translation from subgroup accuracy to group fairness is partly by construction. In Stage~(4) of \S\ref{sec:mechanism}, each member applies $\delta_y$ only to inputs whose true class is $y$, so improvements on the targeted subgroup naturally affect group-level fairness metrics. To check that the perturbation is not equally effective as a generic misclassification tool, Appendix~\ref{app:cross-label} evaluates cross-target use, where the same $\delta_y$ is applied to inputs whose true class is not $y$. The corrective rate substantially exceeds the cross-target rate at every pool size, indicating that the perturbation is a stronger lever for correction than for misuse.

\section{Discussion}
\label{sec:discussion}
\paragraph{Platform-side defenses.} In this work, we assume that platform model is a static black-box classifier and the collective optimizes their perturbation on the query-response pair from this model. However, a platform may also deploy adversarial defenses such as \texttt{Adaptive Randomized Smoothing}~\citep{lyuAdaptiveRandomizedSmoothing} or \texttt{Nasty Teacher}~\citep{maUndistillableMakingNasty2020} to reduce the effectiveness of universal perturbations on the target model or make training of proxy models more difficult. Furthermore, the intervention discussed in this work focuses on users encountering quality-of-service harms. A platform has limited incentive to deploy defenses against this kind of intervention because users' motivation is corrective rather than adversarial, and the platform would presumably prefer corrected predictions. However, the incentives may differ in the case of allocative harms, where the disparity arises from the distribution of a limited resource. Appendix~\ref{app:platform-defenses} shows that platform-side defenses can be effective in preventing TTCA but also cost the platform in terms of accuracy degradation. Whether platform-side robustness disproportionately blocks corrective recourse compared to malicious use is an open empirical question.

\paragraph{Conservative individual baseline.} We compare collective action against targeted SimBA (pixel variant), a simple query-only attack chosen to represent the individual baseline. More query-efficient per-image black-box~\citep{fengGRAPHITEGeneratingAutomatic2022, andriushchenkoSquareAttackQueryEfficient2020, HopSkipJumpAttackQueryEfficientDecisionBased} could reach individual success in fewer queries, which would reduce the cost gap we report. We expect the collective to remain cheaper in aggregate because it pays a fixed proxy-extraction and perturbation-generation cost once, amortizes that cost across the subgroup, and allows non-contributing users to benefit from the resulting perturbation. However, stronger individual attacks would shift the break-even $K^\star$ upward and tighten the per-contributor cost threshold $\lceil N/\overline q\rceil$. Appendix~\ref{app:square-attack}. shows experiments for Square Attack~\citep{andriushchenkoSquareAttackQueryEfficient2020} where the break-even point is higher than SimBA baseline. The Individual vs.\ Collective comparison in \S\ref{sec:results-subgroup} should therefore be interpreted as a conservative estimate of the collective's cost advantage, not as its expected magnitude under the strongest possible individual attack.

\paragraph{Stronger adversarial mechanisms.} We propose one specific way of implementing a test-time collective action framework where we use a KnockoffNets ensemble with a per-class optimized perturbation. Other adversarial methods could be substituted at any stage of the pipeline, both in how the proxy is extracted from the platform and in how the perturbation is optimized against it. Evaluating these systematically is a natural extension and would sharpen the practical guidance the framework can offer a real collective.

\paragraph{Beyond quality-of-service framings.} Our framing treats subgroup-accuracy disparities as quality-of-service harms that affected users should be able to correct. This assumption holds only when the underlying task objective is broadly useful to users. However, this may not work when the task itself is the cause of the harm. For instance, face-attribute classification, particularly in automated
gender recognition, has been criticized in various works. \citet{Keyes18} argues that by treating gender as a binary and externally visible attribute, the harm of automated gender recognition is structural rather than a matter of performance, leading to systems erasing trans identities. \citet{ScheuermanPB19} also in their evaluations find that facial analysis services underperform on transgender individuals and are unable to classify non-binary genders. \citet{shelbySociotechnicalHarmsAlgorithmic2023} formalize such effects as \emph{representational harms}, including erasure of social groups, alienation of those whose membership the system fails to acknowledge, denial of the opportunity to self-identify, and reification of essentialist categories. \citet{Keyes18} recommends that designers of gendered systems should ask whether the system needs to be gendered at all, and if so, how to gender it in a way that recognizes a wide range of people. Hence, although our experiments show that subgroup-accuracy gaps can be reduced technically in a gender classifier, as in our FairFace results, they do not address the harm of deploying such a classifier in the first place. When the task itself is harmful, user-side action may need to move beyond the corrective framing developed in this paper and instead take forms such as refusal, opt-out, or contestation~\citep{Zong2024}.

\section{Conclusion}
\label{sec:conclusion}
We proposed and evaluated test-time collective action: a mechanism through which a coordinated group of users, relying only on standard API access to a deployed classifier, can improve subgroup accuracy for an under-served group without participating in, or relying on, the platform's training pipeline. We frame this as a form of \emph{corrective intervention}: the platform is assumed to be broadly useful, but uneven in the quality of service it provides, and the central question is whether affected users can close this gap from outside the training loop when platform-side remediation is unavailable or delayed. We answer this by using a proxy-based mechanism where the collective, using only their API access to the model, can generate per-class perturbations. This not only addresses the issue by closing the subgroup-accuracy gaps but also translates into fairness metrics and is orders of magnitude fewer API queries than uncoordinated individual interventions. This work contributes to the ongoing discussion on external, user-driven mechanisms to correct harms. TTCA also acts as a complementary approach to the ACA framework in the test-time setting, where the model is already deployed, showing that collective action remains a viable option even when users cannot influence the training loop or know whether their data was used to train the model.

\section*{Acknowledgements}
The resources used in preparing this research were provided, in part, by the Province of Ontario, the Government of Canada through CIFAR, and companies sponsoring the Vector Institute (\url{www.vectorinstitute.ai/partnerships/}). The authors thank the Digital Research Alliance of Canada for computing resources. Ulrich Aïvodji is supported by an NSERC Discovery grant (RGPIN-2022-04006) and an IVADO's Canada First Research Excellence Fund to develop Robust, Reasoning and Responsible Artificial Intelligence (R$^3$AI) grant (RG-2024-290714). Elliot Creager is supported by an NSERC Discovery grant (RGPIN-2024-05116).
\FloatBarrier
\bibliographystyle{plainnat}
\bibliography{references}

\clearpage
\newpage
\appendix
\section{Appendix}
\vspace{2em}
\subsection{Cross-label transfer} \label{app:cross-label}



\begin{figure}[htbp]
  \centering
  \begin{subfigure}[t]{0.48\linewidth}
    \centering
    \includegraphics[width=\linewidth]{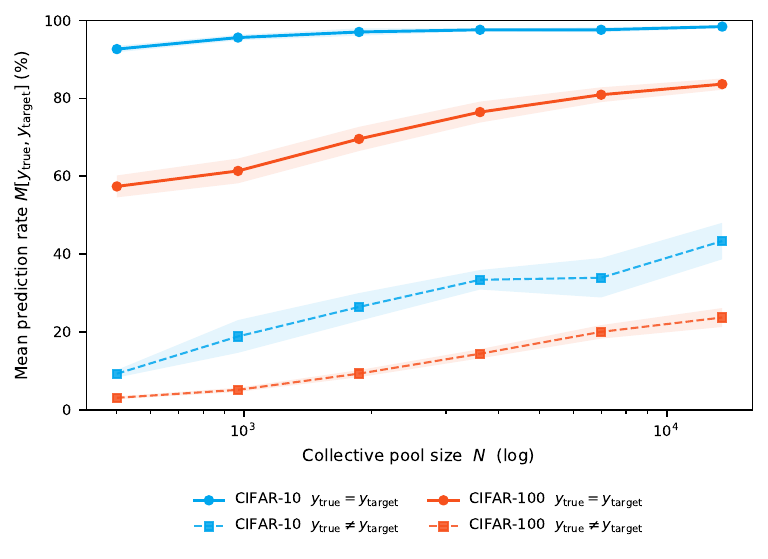}
    \caption{\cifarten and \cifarh.}
    \label{fig:cross-label-cifar}
  \end{subfigure}
  \hfill
  \begin{subfigure}[t]{0.48\linewidth}
    \centering
    \includegraphics[width=\linewidth]{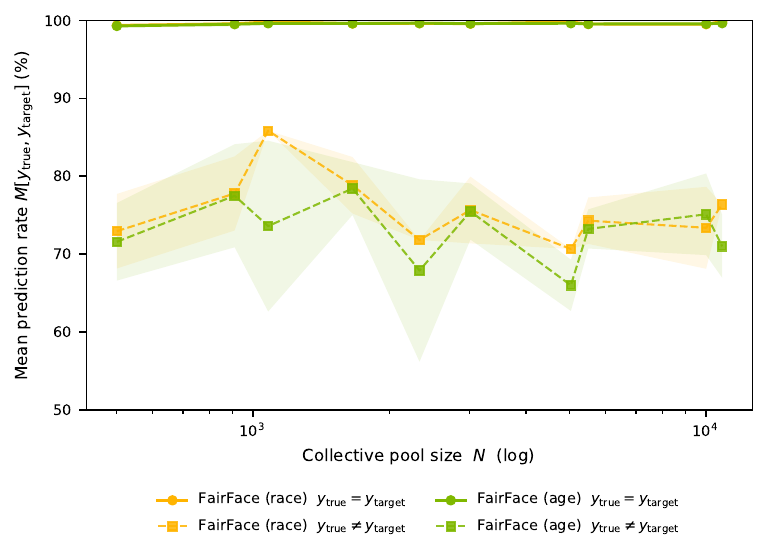}
    \caption{\fairface.}
    \label{fig:cross-label-fairface}
  \end{subfigure}
  \caption{The intervention works best as a corrective tool. Solid: $M[y_{\mathrm{true}} = y_{\mathrm{target}}]$, $\delta_y$ pushes predictions toward the true class $y$ (corrective). Dashed: $M[y_{\mathrm{true}} \neq y_{\mathrm{target}}]$, $\delta_y$ pushes toward a different class (cross-target). The gap shows pooled extraction is a much stronger lever for correction than for cross-target use. Corrective use leads on \fairface too, though the gap is narrower than CIFAR's. Subgroups are bottom-$k$ (gender, race) or (gender, age) cells against a binary gender label, so cross-target $\delta_y$ shares movement across cells.}
  \label{fig:cross-label-transfer}
\end{figure}

The collective produces a per-class perturbation $\delta_y$ for each target class $y$ that appears in $\mathcal{G}$. The intended use is corrective: a member whose true class is also $y$ applies $\delta_y$ to push the platform's prediction toward the correct label. Once extracted, however, the same $\delta_y$ can be applied to inputs whose true class is not $y$, where it pushes the prediction toward $y$ regardless of the input's ground truth. Figure~\ref{fig:cross-label-transfer} plots both rates (prediction rate of $y$ when true class is $y$ and when true class is not $y$) as a function of pool size $Q$ across CIFAR-10, CIFAR-100, FairFace (race), and FairFace (age).

On CIFAR-10 and CIFAR-100, the corrective rate significantly exceeds the cross-target rate at every pool size. An attacker hoping to misclassify an input would achieve a much lower success rate using $\delta_y$ than using a dedicated attack, so the perturbation is a stronger lever for correction than for misuse on these benchmarks. The FairFace panels show a much narrower gap: cross-target rates of 65 -85\% compared to the corrective rate of ${\sim}100\%$. This may be because $M$ is computed at the binary gender label while the subgroup is defined as (gender, race) or (gender, age) pair, so a subgroup-specific $\delta_y$ moves the gender prediction in a way that is shared across multiple pairs with the same gender label. 

\subsection{Platform defenses}
\label{app:platform-defenses}

\begin{figure}[t]
  \centering
  \includegraphics[width=0.80\columnwidth]{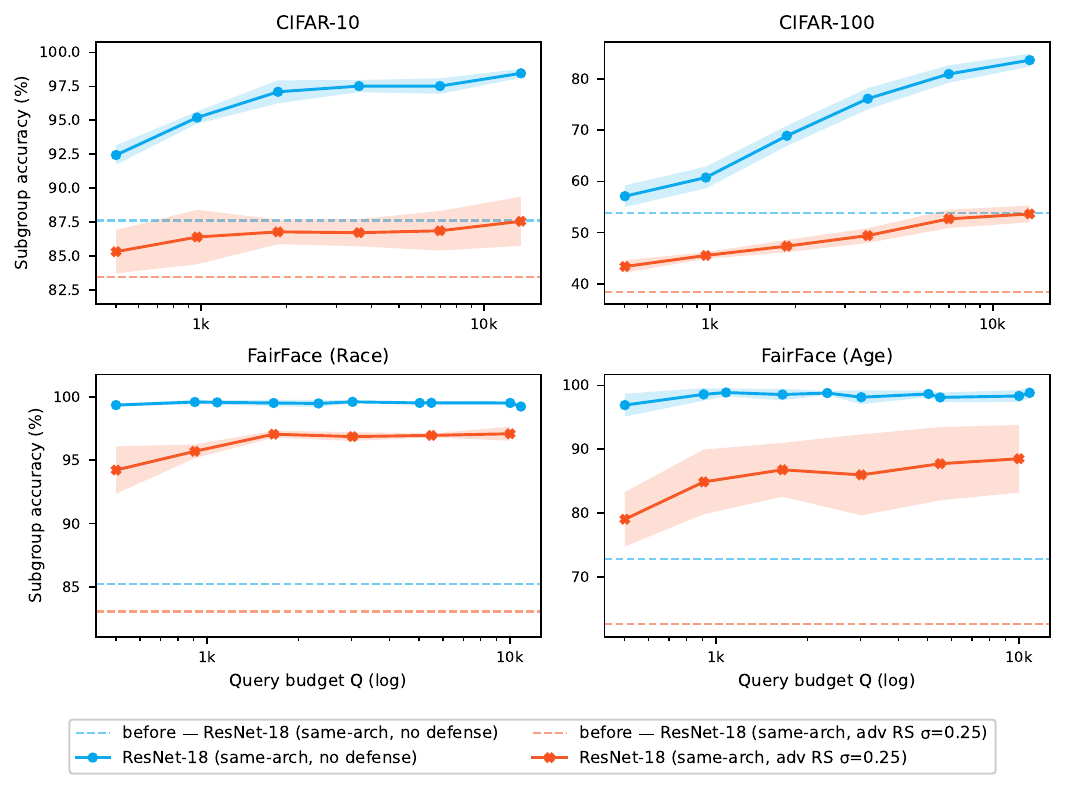}
  \caption{\textit{Subgroup accuracy after $\delta_y$ with and without randomized smoothing.} Solid lines show subgroup accuracy after applying $\delta_y$ for an undefended ResNet-18 (blue) and one trained with adversarial randomized smoothing at $\sigma=0.25$ (red); dashed lines show the pre-perturbation baseline. The defense reduces TTCA's corrective gain but does not eliminate it, and itself lowers the platform's pre-perturbation accuracy on the targeted subgroup. On FairFace the defended classifier after $\delta_y$ still exceeds the undefended baseline, while on CIFAR-10 it barely matches it.}
  \label{fig:platform_defenses}
\end{figure}

To understand how TTCA would perform under conditions where the platform employs defenses, we compare the subgroup accuracy on \cifarten and \cifarh on a \resneteighteen classifier with Randomized Smoothening as shown in Figure.~\ref{fig:platform_defenses}.  The corrective effect is present across all four benchmarks but is comparatively lower after randomized smoothening. However, the defense imposes a cost on the platform itself. Randomized smoothing lowers pre-perturbation subgroup accuracy on the targeted group across all benchmarks (dashed red below dashed blue), so the platform's choice to defend against corrective TTCA trades resistance for accuracy on the very subgroup TTCA is designed to help. This is also discussed in Platform-side defenses in \S\ref{sec:discussion}.

\subsection{Probing Perturbations at the lowest \texorpdfstring{$Q$}{Q}} \label{app:white-image-test}



\begin{figure}[htbp]
    \centering
    \begin{subfigure}[c]{0.32\linewidth}
        \centering
        \includegraphics[width=\linewidth]{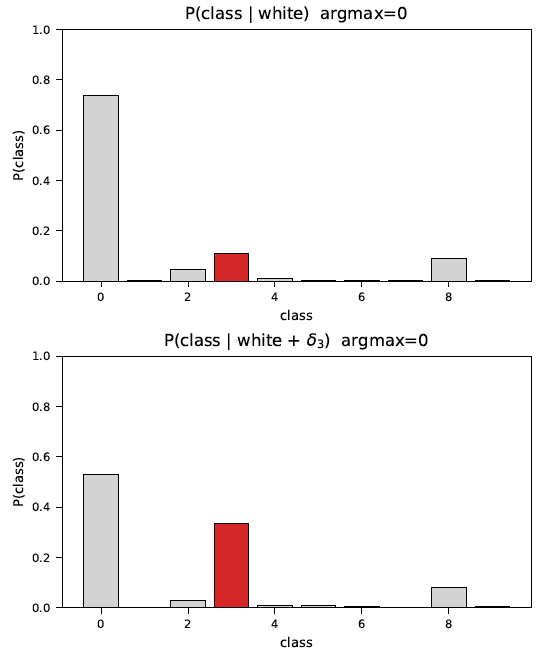}
        \caption{White canvas test.}
        \label{fig:white-canvas}
    \end{subfigure}
    \hfill
    \begin{subfigure}[c]{0.62\linewidth}
        \centering
        \includegraphics[width=\linewidth]{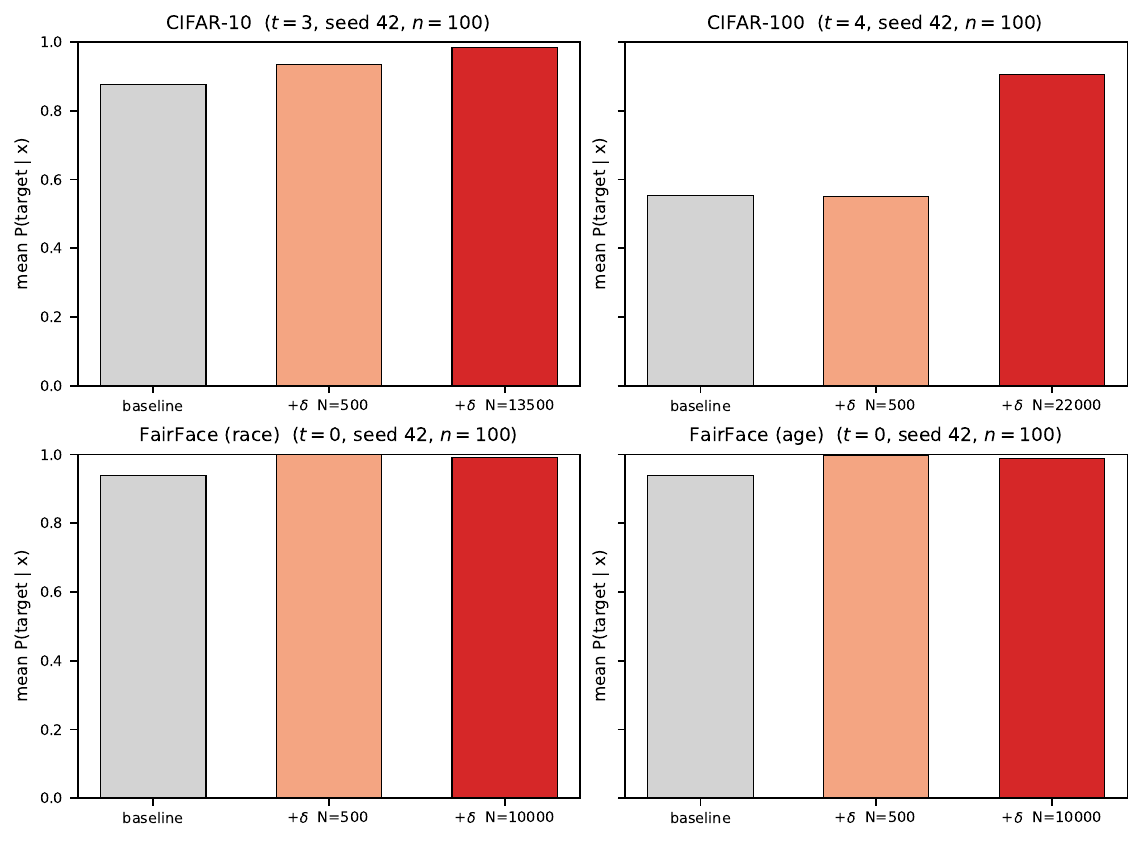}
        \caption{Previously misclassified image.}
        \label{fig:misclassified}
    \end{subfigure}
    \caption{\textbf{Left:} To isolate what $\delta_y$ encodes from any image content it might be exploiting, we apply it to a white canvas on \cifarten where the worst performing class was \texttt{class=3}. Even at ($Q=500$), the teacher's predicted probability of the target class $y$ on this $\text{white canvas} + \delta_y$ input goes up. \textbf{Right:} To further see how the class prediction probabilities change on a previously misclassified image, we sample a random misclassified image and apply the perturbation at $Q=500$ and highest $Q$ for the dataset. For every dataset except \cifarh, the delta at $Q=500$ increases the class probability.}
    \label{fig:placeholder}
\end{figure}

The two panels of Figure~\ref{fig:placeholder} probe what the
perturbation actually learns at the smallest pool size we
evaluate. Applied on its own to an input with no image content (white canvas), $\delta_y$ already raises the platform's predicted probability of class $y$. When applied to real subgroup members the platform misclassifies, the same $\delta_y$ at $Q=500$ raises the predicted probability of the true class across every benchmark except \cifarh. Together these observations show the proxy does not need to replicate the platform's behavior everywhere to be useful but it only needs to recover enough of the decision boundary near class $y$. A small pool is already enough for that local recovery, which is why subgroup accuracy lifts at $Q=500$ in \S\ref{sec:results-subgroup} without the proxy's overall fidelity being high.

\FloatBarrier

\subsection{Square Attack as Individual baseline}
\label{app:square-attack}

Square Attack is a black-box attack that uses randomized search to find localized square-shaped updates at random positions to optimally generate adversarial examples~\citep{andriushchenkoSquareAttackQueryEfficient2020}. The assumption of the attack aligns with the individual baseline used in \S\ref{sec:ind-vs-coll} (\emph{i.e.,} black-box access with probability vectors). As previously mentioned in \S\ref{sec:discussion}, a query-efficient individual attack may reduce the magnitude of gain of the collective over individuals attacking alone. In this experiment, we adapt the implementation of Square Attack from \texttt{Adversarial Robustness Toolbox}~\citep{nicolaeAdversarialRobustnessToolbox2019} for a targeted attack variant. Additionally, we also add a counter on top of the predict function of the classifier to count the number of platform queries made. Table~\ref{tab:individual-vs-collective-cifar100-square} shows the results from the experiment, showing that the collective advantage decreases when compared to SimBA. However, for B=500, the collective can still beat the individual attack with 37 members. An important note here is that the original work does highlight that the targeted attack variant may be less query efficient on margin-based loss, as used in this experiment, and a cross-entropy loss can further make the individual attacking baseline stronger.

\begin{table}[t]
  \centering
  \caption{Query-budget analysis on CIFAR-100 (ResNet-18 platform and proxy). At every per-user budget $B$ below the collective\textquotesingle s $\sim$83\% accuracy ceiling, pooling matches the individual Square baseline with far fewer platform queries, and the margin grows as $B$ is higher. The highlighted row is the largest $B$ at which a collective of $K^\star$ members still matches individual; the rows below it are budgets where individual surpasses the collective ceiling, and no pool size, including the full $\mathcal{D}_{\mathrm{pub}}$, closes the gap.}
  \label{tab:individual-vs-collective-cifar100-square}
   \scriptsize
  \setlength{\tabcolsep}{1.4pt}
  \renewcommand{\arraystretch}{1.0}
  \begin{tabular}{@{}ccccccc@{}}
    \toprule
      \makecell[c]{Query \\Budget\\($B$)} &
      \makecell[c]{Individual\\acc.\,(\%)} &
      \makecell[c]{Collective\\acc.\,(\%)} &
      \makecell[c]{Total\\ Collective\\Query Pool ($N$)} &
      \makecell[c]{Collective\\Members\\ ($K^\star$)} &
      \makecell[c]{Avg. \\Individual\\queries ($\overline q$)} &
      \makecell[c]{Total \\Individual\\queries ($\sum q_i$)} \\
    \midrule
    100 & 79.4{\tiny\,\textpm\,1.4} & 81.7{\tiny\,\textpm\,2.0} & 8{,}286{\tiny\,\textpm\,2{,}914} & 83 & 74{\tiny\,\textpm\,4} & 74{,}332{\tiny\,\textpm\,4{,}030} \\
    200 & 81.1{\tiny\,\textpm\,1.8} & 82.9{\tiny\,\textpm\,2.1} & 10{,}893{\tiny\,\textpm\,3{,}570} & 55 & 132{\tiny\,\textpm\,9} & 132{,}449{\tiny\,\textpm\,9{,}111} \\
    \rowcolor{green!15!gray!10} \textbf{500} & \textbf{84.4{\tiny\,\textpm\,1.8}} & \textbf{83.3{\tiny\,\textpm\,2.2}} & \textbf{18{,}600{\tiny\,\textpm\,4{,}656}} & \textbf{37} & \textbf{271{\tiny\,\textpm\,27}} & \textbf{271{,}247{\tiny\,\textpm\,27{,}036}} \\
    \midrule
      1{,}000 & 87.4{\tiny\,\textpm\,1.5} & 83.0{\tiny\,\textpm\,2.6} & 22{,}000{\tiny\,\textpm\,0} & 22 & 443{\tiny\,\textpm\,47} & 442{,}552{\tiny\,\textpm\,47{,}330} \\
      2{,}000 & 89.2{\tiny\,\textpm\,2.1} & 83.0{\tiny\,\textpm\,2.6} & 22{,}000{\tiny\,\textpm\,0} & 11 & 752{\tiny\,\textpm\,126} & 751{,}788{\tiny\,\textpm\,126{,}351} \\
      5{,}000 & 91.8{\tiny\,\textpm\,1.3} & 83.0{\tiny\,\textpm\,2.6} & 22{,}000{\tiny\,\textpm\,0} & 5 & 1{,}400{\tiny\,\textpm\,232} & 1{,}400{,}092{\tiny\,\textpm\,231{,}769} \\
      10{,}000 & 93.7{\tiny\,\textpm\,1.0} & 82.3{\tiny\,\textpm\,2.3} & 22{,}000{\tiny\,\textpm\,0} & 3 & 2{,}129{\tiny\,\textpm\,332} & 2{,}128{,}572{\tiny\,\textpm\,332{,}311} \\
    \bottomrule
  \end{tabular}
\end{table}
\FloatBarrier

\end{document}